\documentclass{article}
\pdfoutput=1

\usepackage{arxiv}

\usepackage[utf8]{inputenc} 
\usepackage[T1]{fontenc}    
\usepackage{hyperref}       
\usepackage{url}            
\usepackage{booktabs}       
\usepackage{amsfonts}       
\usepackage{nicefrac}       
\usepackage{microtype}      
\usepackage{lipsum}
\usepackage{subfig}
\usepackage{graphicx}
\graphicspath{ {./images/} }
\usepackage{amsmath,amssymb}
\usepackage{geometry}
\usepackage{multirow, booktabs}
\usepackage{makecell}

\title{The Data Representativeness Criterion:\\ Predicting the Performance of Supervised Classification Based on Data Set Similarity}

\author{
 Evelien Schat \\
  Department of Methodology and Statistics \\
  Utrecht University\\
  Padualaan 14, Utrecht, The Netherlands\\
  and \\
  Netherlands eScience Center \\
  Science Park 140, Amsterdam, The Netherlands \\
  \texttt{evelien.schat@gmail.com} \\
   \And
 Rens van de Schoot \\
  Department of Methodology and Statistics \\
  Utrecht University\\
  Padualaan 14, Utrecht, The Netherlands\\
  and \\
  Optentia Research Focus Area \\
  North-West University \\
  Vanderbijlpark 1900, South Africa
  \And
 Wouter M. Kouw \\
  Netherlands eScience Center \\
  Science Park 140, Amsterdam, The Netherlands \\
  \And
 Duco Veen \\
  Department of Methodology and Statistics \\
  Utrecht University\\
  Padualaan 14, Utrecht, The Netherlands\\
  \And
 Adri{\"e}nne M. Mendrik \\
  Netherlands eScience Center \\
  Science Park 140, Amsterdam, The Netherlands
}

\begin{document}
\maketitle
\begin{abstract}
In a broad range of fields it may be desirable to reuse a supervised classification algorithm and apply it to a new data set. However, generalization of such an algorithm and thus achieving a similar classification performance is only possible when the training data used to build the algorithm is similar to new unseen data one wishes to apply it to. It is often unknown in advance how an algorithm will perform on new unseen data, being a crucial reason for not deploying an algorithm at all. Therefore, tools are needed to measure the similarity of data sets. In this paper, we propose the Data Representativeness Criterion (DRC) to determine how representative a training data set is of a new unseen data set. We present a proof of principle, to see whether the DRC can quantify the similarity of data sets and whether the DRC relates to the performance of a supervised classification algorithm. We compared a number of magnetic resonance imaging (MRI) data sets, ranging from subtle to severe difference is acquisition parameters. Results indicate that, based on the similarity of data sets, the DRC is able to give an indication as to when the performance of a supervised classifier decreases. The strictness of the DRC can be set by the user, depending on what one considers to be an acceptable underperformance. 

\end{abstract}


\section{Introduction}
Generalization of supervised classification algorithms to new unseen data sets, is limited to the data set’s similarity to the available training data. It is unclear in advance whether an algorithm will perform well on unseen data, which is a critical reason for not deploying an algorithm. In order to get an indication of the algorithm's performance on the unseen data, it is essential to develop tools that measure representativeness. This becomes essential in the more subtle cases, where it is hard for humans to predict whether algorithms will have a similar performance on the unseen data as on the training data. An example of this is brain tissue classification in magnetic resonance imaging (MRI) data. MRI scans acquired with different protocols, may seem similar to the human eye (human vision), but can have drastic influence on the performance of automatic brain tissue classification algorithms (computer vision) \cite{van2014transfer}. 

In this paper, we introduce the Data Representativeness Criterion (DRC) to predict the generalization of a supervised classification algorithm to new unseen data. After determining the distribution overlap between the training data and the new unseen data, the DRC could be used to predict generalization, without the need for labelled data. With the DRC, we aim to determine the threshold \textit{when} additional actions are required in order to improve classification performance on unseen data. These actions could exist of labeling part of the unseen data, such that it could be used for retraining a supervised machine learning algorithm (e.g. active learning \cite{cohn1995active, ghasemi2011active}), to quickly generalize to the unseen data. Or by using methods such as data augmentation \cite{van2001art}, transfer learning \cite{pan2010survey, kouw2019, van2014transfer} or representation learning \cite{bengio2013representation,kouw2019learning}, which are commonly used to extend the scope of machine learning algorithms.

The DRC is based on Bousquet’s Data Agreement Criterion (DAC) \cite{bousquet2008diagnostics}, but has been adjusted to assess data set similarity. The idea of assessing data set similarity is based on the work described in \cite{kouw2019learning}. In this paper, the proxy ${\cal A}$-distance was introduced as an approximation to data set similarity in the context of MRI data sets to evaluate representation learning. We combined aspects of the proxy ${\cal A}$-distance and the DAC, resulting in the DRC measure. Both the proxy ${\cal A}$-distance and the DRC are based on the similarity between the training and unseen data sets. Section \ref{Method} first describes how the data set similarity is determined, after which the proxy ${\cal A}$-distance is described and the DRC is introduced. Section \ref{Experiments} describes a controlled experiment, to show how the DRC behaves with different benchmark priors. Based on brain tissue segmentations of real human brain data, we obtained a number of different MRI data sets, ranging from subtle to severe differences in protocol (acquisition parameters). Both the proxy ${\cal A}$-distance and the DRC are applied to this data, to show how they relate to the supervised tissue classification performance. In Section \ref{results} the results of the study are presented, followed by a discussion and conclusion in Sections \ref{discussion} and \ref{conclusion}. The data and Python code of the controlled experiment are available at https://github.com/eschat/DRC.

\section{Methods}\label{Method} 
In this section we elaborate on data set similarity and provide a description of the proxy ${\cal A}$-distance, DAC and DRC. Moreover, we provide a rationale of how aspects of the proxy ${\cal A}$-distance and DAC are combined, resulting in the DRC measure.

\subsection{Data Set Similarity}\label{DataSetSimilarity}
To predict the performance of a supervised classifier, it is first necessary to establish the similarity of the data sets in question based on their underlying distributions. To establish the similarity, we depend on the ability of a classifier to discriminate between domains: training data from domain T and unseen data from domain U.


In situations where two data sets are very similar, there is a large amount of overlap between the underlying distributions of domains T and U. Classification probabilities are thus expected to be around 0.5, as the domain classifier will have difficulties distinguishing between the domains. As the difference between two data sets increases, there will be less overlap between the underlying distributions. The further apart the two data sets, the more the classification probabilities are expected to shift towards 0 or 1, indicating that the classifier has less difficulties distinguishing between the domains.

\subsection{Proxy\texorpdfstring{$\cal{A}$}{Lg}-distance}\label{ProxyA}
The proxy ${\cal A}$-distance \cite{ben2007analysis,ben2010theory}, denoted by $d_{\cal A}$, is an empirical distance measure between two data sets and depends on the ability of a classifier to discriminate between domain T and domain U. The measure is derived from the more general \emph{total variation} distance, which can be thought of as the largest difference between probabilities $x$ assigned by probability distributions $p$ and $q$ to the same event. This distance however cannot be computed, therefore two steps are taken to approximate it. Firstly, the distance can be rewritten to $\ 2 \big( 1 - \int \min \{p(x), q(x)\} \mathrm{d}x \big)$, provided the sample space is countable \cite{ben2007analysis}. Secondly, in $\int \min \{p(x), q(x)\} \ \mathrm{d}x$, one recognizes the error of a classification function that discriminates between the two distributions $e(p,q)$. The distance can be approximated using samples from two data sets:
\begin{align} \label{eq:PAD}
	d_{\cal A}(S_T,S_U) = 2\big(1 - 2 \ \hat{e}[S_T,S_U] \big) \, ,
\end{align}
where $\hat{e}$ refers to the cross-validation error between data set $S_T$ (training) and data set $S_U$ (unseen). This distance $d_{\cal A}$ is referred to as the proxy-${\cal A}$ distance \cite{ben2010theory}. A test error of 0 corresponds to a proxy ${\cal A}$-distance of 2. This means that the training and unseen data are perfectly separable. A test error of 0.5 corresponds to a proxy ${\cal A}$-distance of 0. In this case, the training and unseen data sets cannot be distinguished. The lower the proxy ${\cal A}$-distance, the more similar the training and unseen data. 

The proxy-${\cal A}$ distance suffers from a limitation common to many other distance measures: how should the quantitative value, lying in the interval $[0, 2]$, be interpreted to the qualitative value $\{"\text{similar}", "\text{dissimilar}"\}$? It is clear that a threshold on distance is required before data set similarity can be considered. In the following, we combine aspects from the proxy-${\cal A}$ distance with the Data Agreement Criterion and a set of reference priors to form interpretable thresholds.

\subsection{Data Representativeness Criterion}\label{DRC}
The DAC is a measure of prior-data conflict \cite{bousquet2008diagnostics} and has been used to evaluate expert knowledge (i.e. prior information) in light of new data \cite{veen2018using,schalken2018dac}. Taking into account the proxy-${\cal A}$ distance’s limitation of having no clearly defined threshold, we adapted the DAC to fit the context of comparing data sets, resulting in the DRC measure. 

The DRC is based on a ratio of Kullback-Leibler (KL) divergences \cite{kullback1951information}. A KL divergence is a measure of informative regret and measures the information lost when a distribution $\pi_2(\theta)$ is used to approximate a reference distribution $\pi_1(\theta)$. The larger the KL divergence, the larger the difference between the two distributions in question. Following the definition offered by Bousquet \cite{bousquet2008diagnostics}, the KL divergence between distributions $\pi_1(\theta)$ and $\pi_2(\theta)$ is as follows:

\begin{align} \label{eq:KL}
	KL(\pi_1||\pi_2) = \int_{\Theta} \pi_1(\theta) \log\frac{\pi_1(\theta)}{\pi_2(\theta)}\mathrm{d}\theta,  \, 
\end{align}   

\noindent
where $\Theta$ denotes the set of all values for the parameter $\theta$, $\pi_1(\theta)$ denotes the reference distribution and $\pi_2(\theta)$ denotes the approximating distribution. Using the KL divergence as in Eq \ref{eq:KL}, the DRC is defined as:

\begin{align} \label{eq:DRC}
  \text{DRC} = \dfrac{KL[\pi_{TU}(\theta)||\pi_{bm1}(\theta)]}{KL[\pi_{TU}(\theta)||\pi_{bm2}(\theta)]}, \, 
\end{align}

\noindent
where $\pi_{TU}(\theta)$ denotes the distribution representing the separability of the training data $S_T$ and unseen data $S_U$. The distribution is based on classification probabilities of a domain classifier, build to distinguish between the training data and unseen data. Furthermore, $\theta$ represents the classification probabilities and $\pi_{bm1}(\theta)$ and $\pi_{bm2}(\theta)$ denote benchmark prior 1 and benchmark prior 2, respectively. Benchmark prior 1 represents the separability distribution of two similar data sets while benchmark prior 2 represents the separability distribution of two dissimilar data sets. 

As we are comparing 2 domains, beta distributions are used for $\pi_{TU}(\theta)$ and the benchmark priors. Note that in situations where there are more than 2 domains, a Dirichlet distribution can be used. By definition, if the DRC is smaller than 1, $\pi_{TU}(\theta)$ and $\pi_{bm1}(\theta)$ resemble each other more closely than $\pi_{TU}(\theta)$ and $\pi_{bm2}(\theta)$. If the DRC is larger than 1, more information is lost when choosing benchmark prior 1 as compared to benchmark prior 2. 

The DRC is based on classification probabilities, meaning that the measure is a probabilistic one. This is in contrast with the proxy ${\cal A}$-distance, which is a deterministic measure as it does not take uncertainty in classification into account. The proxy ${\cal A}$-distance merely looks at the most likely class (i.e. domain). 

\subsubsection{Determining Benchmark Priors} \label{DRC_bm}
As we want to compare the separability (i.e. dissimilarity) of different data sets, the data is the variable of interest. This leads to the separability distribution $\pi_{TU}(\theta)$ being the dynamic component in the DRC. To be able to compare these separabilities, the benchmark priors are fixed points of reference. This is unlike the original DAC, where the prior information based on an expert is the variable of interest and the data is the fixed point of reference.

The benchmark priors are chosen such that a DRC larger than 1 indicates that the training and unseen data are not exchangeable (i.e. algorithm will under-perform when applied to the unseen data). Consequently, a DRC smaller than 1 indicates that the training data is representative of the unseen data (i.e. algorithm will have a similar performance when applied to the unseen data). 
A DRC is smaller than 1 when $\pi_{TU}(\theta)$ is more similar to benchmark prior 1 than benchmark prior 2. Benchmark prior 1 represents the separability of two \textit{similar} data sets. On the other hand, a DRC is larger than 1 when $\pi_{TU}(\theta)$ is more similar to benchmark prior 2 than benchmark prior 1. Benchmark prior 2 thus represents the separability of two \textit{dissimilar} data sets. A DRC of 1 is a special case, where the separability distribution is as similar to benchmark prior 1 as to benchmark prior 2.

Ideally, benchmark prior 2 should be a distribution which represents two data sets that are completely separable. Two completely separable data sets would result in classification probabilities around 0 and 1, which in turn would lead to an improper beta distribution. The DRC requires distributions to be proper \cite{bousquet2008diagnostics} and therefore, we set benchmark prior 2 as a Beta(1, 1) distribution. We argue that two data sets do not need to be completely separable before we can determine that one is not representative of another and that generalization of an algorithm is not possible. As such, a Beta(1, 1) distribution would already be a suitable worst-case scenario. In Section \ref{DRC_parameters}, we elaborate on the shape parameters of benchmark prior 1.

\section{Experiments}\label{Experiments}

\begin{figure}
    \centering
    \includegraphics[width=\textwidth]{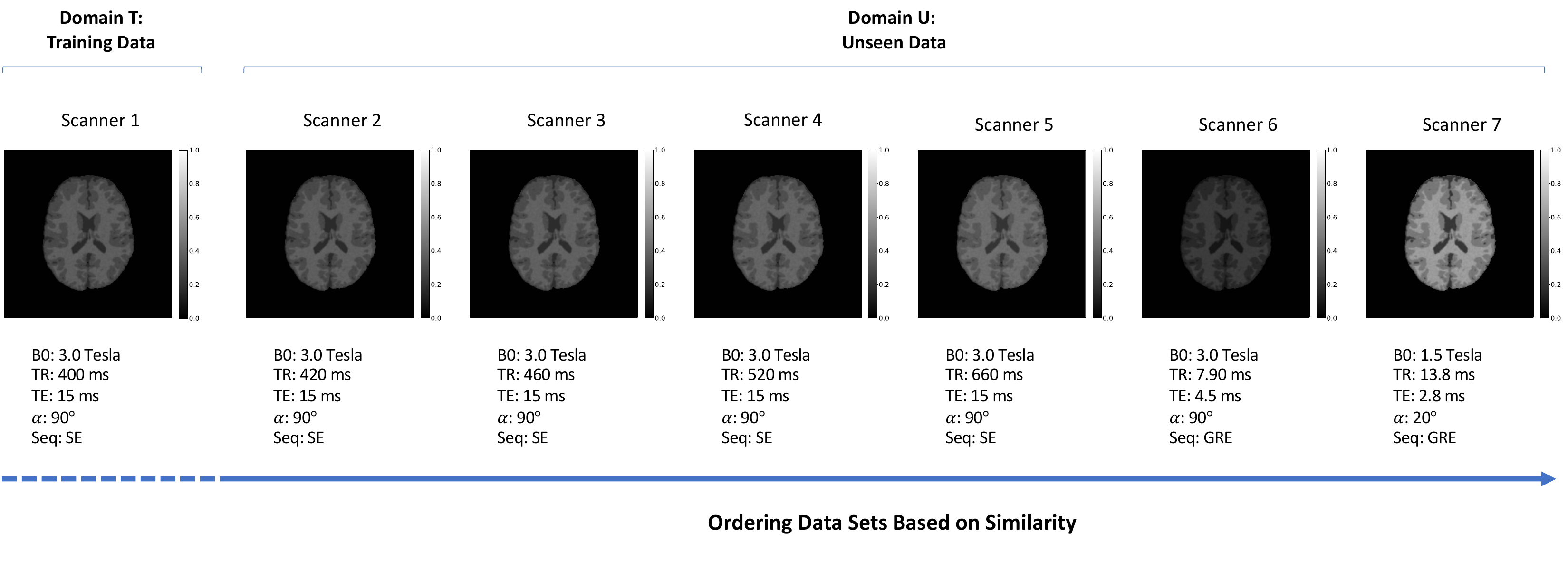}
    \caption{Examples of segmentations with corresponding acquisition parameter settings. In the controlled experiment, we compared scanner 1 (domain T: training data) with the other 6 scanners (domain U: unseen data). The difference between scanner 1 and the additional 6 scanners ranged from subtle to severe differences in acquisition parameters. The arrow gives an indication of the ordering of the data sets based on similarity, as compared to scanner 1.}
    \label{fig:segmentations}
\end{figure}

We present a controlled experiment, to see whether the DRC could be used to predict supervised classification performance on new unseen data. In the controlled experiment, domain classification was performed to determine whether there is overlap between the distribution of the training data and the distribution of the unseen data (i.e. overlap between domains T and U). Using the output of the domain classifier, we obtained the DRC and proxy ${\cal A}$-distance.

Based on brain tissue segmentations of real human brain data, we obtained a number of different MRI data sets, ranging from subtle to severe differences in protocol (acquisition parameters). Figure \ref{fig:segmentations} shows examples of segmentations with corresponding acquisition parameter settings. More information regarding the data and parameter settings can be found in Section \ref{Data}.

In each condition of the controlled experiment, the two domains were specified. Specifically, scanner 1 (domain T: training data) was compared with the other 6 scanners (domain U: unseen data). In condition 1, we compared scanner 1 with scanner 2, with only a very small difference in acquisition parameters (i.e. small difference in TR). In the following three conditions, scanner 1 was compared with scanners 3, 4 and 5, respectively. The difference in acquisition parameters increased with each condition, by an increase in the value for TR. In condition 5, we compared scanner 1 with scanner 6, both 3.0 Tesla scanners but with very different acquisitions parameters. Lastly, condition 6 compared scanners 1 and 7. Here we compared scanners with different magnetic field strengths: a 3.0 Tesla scanner with a 1.5 Tesla scanner. 

\subsection{Data} \label{Data}
Using segmentations based on real human brain data, we obtained a number of different MRI data sets by simulating the acquisition of the scans. This was done using an MRI simulator \cite{benoit2005simri}, where anatomical models of the human brain were used as input. The anatomical models have been obtained from Brainweb\footnote{http://www.bic.mni.mcgill.ca/brainweb/} and consist of transverse slices of 20 subjects with a normal, healthy brain \cite{aubert2006twenty,aubert2006new,collins1998design}.

Figure \ref{fig:segmentations} shows the acquisition parameters of the different data sets: magnetic field (B0), repetition time (TR), echo time (TE), flip angle ($\alpha$) and sequence (Seq). Each data set represented a scanner. The parameters of the first five scanners were based on optimal scan parameters and adjustable ranges for T1-weighted 3.0 Tesla scanners \cite{lu2005}. We only varied TR, as the adjustable ranges are based on TR and the other parameters are fixed. Scanner 6 was based on a standard protocol for a 3.0 Tesla scanner \cite{mendrik2015mrbrains} and scanner 7 on a standard protocol for a 1.5 Tesla scanner \cite{ikram2015rotterdam}. The arrow in Figure \ref{fig:segmentations} gives an indication of the ordering on the data sets based on similarity, as compared to scanner 1. 
For each scanner, we obtained 20 T1-weighted MRI scans. The images were 256 by 256 pixels, with a 1.0x1.0 mm resolution. We normalized the grey-scale values and used a brain mask to strip the skull. The intensity values in MRI scans are relative and not absolute values (unlike values such as Hounsfield units in CT images). 

The MRI scans were decomposed into patches of 15 by 15 pixels. To limit the influence of the background pixels on classification, all patches in which the middle pixel contained background information were filtered out. Background pixels are not important for classification, as the background contains no information regarding the separability of different MRI data sets. 

\subsection{Data Set Similarity}
As mentioned above, domain classification was performed to determine the similarity of the training and unseen data sets. A logistic regression classifier was used, which was $\ell_2$-regularized and cross-validated for optimal regularization parameters. The domain classifier was built using both training and unseen data, with corresponding domain label. The domain classifier was then tested on both training and unseen data. Specifically, 15 scans from domain T and 15 scans from domain U (100-5,000 random patches per scan) were used for building the domain classifier. 5 scans from domain T and 5 scans from domain U (100-5,000 random patches per scan) were used for testing the domain classifier. For each condition, domain classification was repeated 50 times, due to random sampling of patches.

The domain classification error was used to obtain the proxy ${\cal A}$-distance, as defined in Eq \ref{eq:PAD}. Additionally, the classification probabilities were used for the DRC. For each test patch, two probabilities were given: one probability for it belonging to domain T and one for it belonging to domain U. A beta density function was fitted on all these probabilities taken together. This density function, together with the benchmark priors, were used to obtain the DRC as defined in Eq \ref{eq:DRC}.

\subsection{DRC Parameters} \label{DRC_parameters}

\begin{figure}
    \centering
    \includegraphics[width=0.6\textwidth]{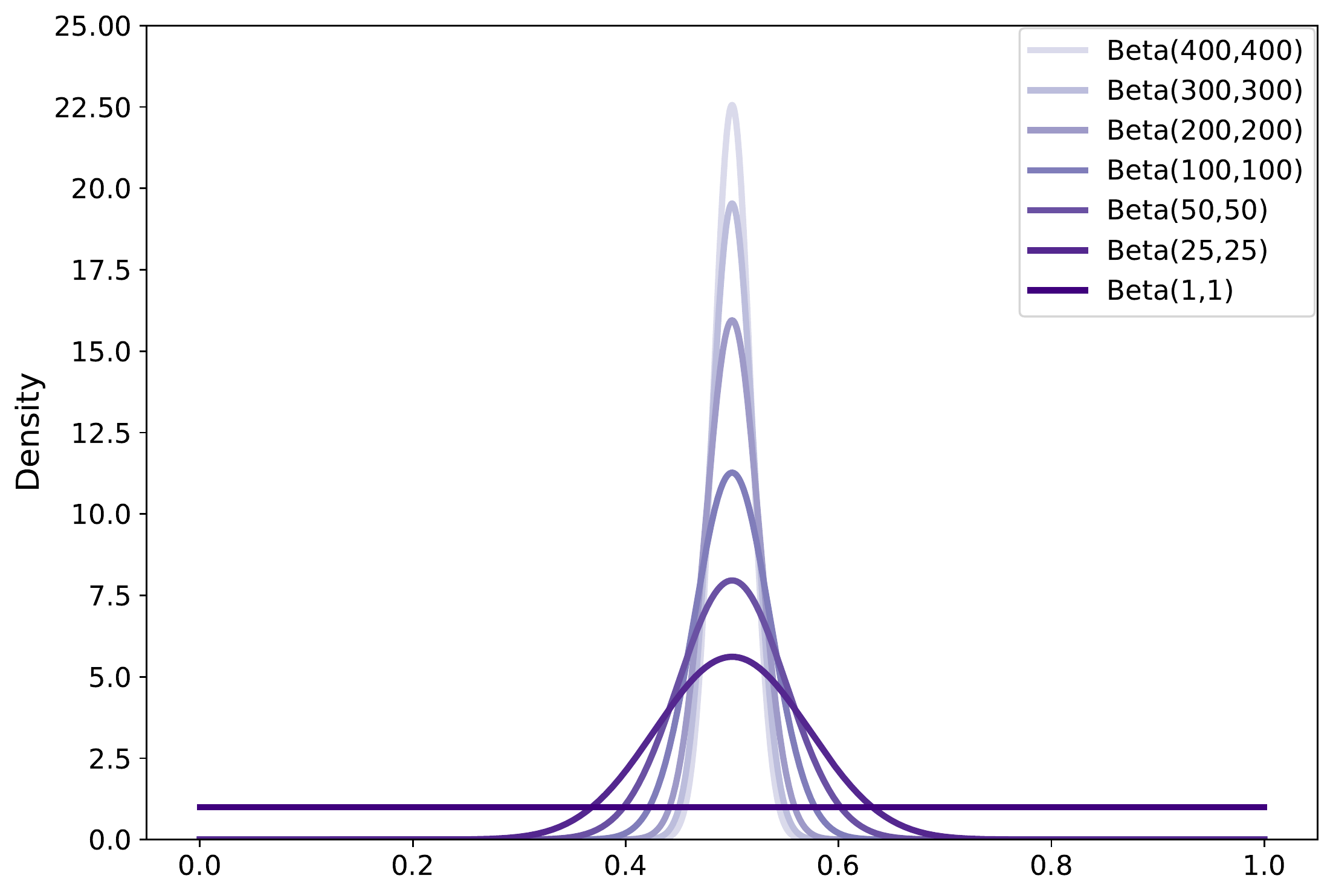}
    \caption{Different benchmark prior distributions for the DRC.}
    \label{fig:beta_distributions}
\end{figure}

In the controlled experiment, we also looked at how the DRC behaves with different benchmark priors. As discussed in Section \ref{DRC_bm}, the separability distribution is the dynamic component in the DRC, while the benchmark priors are fixed points of reference. Also recall that benchmark prior 1 represents the separability of two \textit{similar} data sets while benchmark prior 2 represents the separability of two \textit{dissimilar} data sets. We reasoned that a Beta(1, 1) distribution is suitable for benchmark prior 2. For benchmark prior 1, on the other hand, multiple options are possible. In the controlled experiment, the beta shape parameters of benchmark prior 1 were varied, to see how the DRC changes which different distributions for benchmark prior 1. Specifically, the following distributions were used: Beta(25, 25), Beta(50, 50), Beta(100, 100), Beta(200, 200), Beta(300, 300) and Beta(400, 400). Figure \ref{fig:beta_distributions} shows the different benchmark prior distributions. 

\subsection{Tissue Classification}
Tissue classification was performed to test the effect on tissue classification performance when adding samples of the unseen data to the training data. Two classifiers were used: 1) training classifier (training + unseen): a convolution neural network (CNN) built on both training and unseen data and 2) unseen classifier (unseen): a CNN built only on unseen data. The classifiers were built to classify grey matter, white matter and cerebrospinal fluid. Both classifiers were tested on only unseen data.

For the training + unseen classifier, 15 scans from domain T (7,000 random patches per scan) and 5 scans from domain U (varying from 100-18,000 random patches per scan) were used for building the classifier. 15 independent scans from domain U (7,000 random patches per scan) were used for testing the classifier. For the unseen classifier, 5 scans of domain U (varying from 100-18,000 random patches per scan) were used for building the classifier. 15 independent scans from domain U (7,000 random patches per scan) were used for testing the classifier. For each condition, tissue classification was repeated 10 times, due to random sampling of patches. Here we limited the repetitions to 10 times, as the tissue classification was computationally expensive.

Furthermore, we also performed tissue classification to illustrate the effect of building a tissue classifier on training data and applying it to unseen data. For all six conditions, a CNN was built on training data and applied to unseen data. Specifically, the tissue classifier was built using 15 scans from domain T (7,000 random patches per scan) and was then applied to 1 scan from domain U (all patches in scan).

\section{Results} \label{results}

\subsection{The Effect of Data Set Similarity on Tissue Classification}

\begin{figure}
	\centering
	\includegraphics[width=0.9\textwidth]{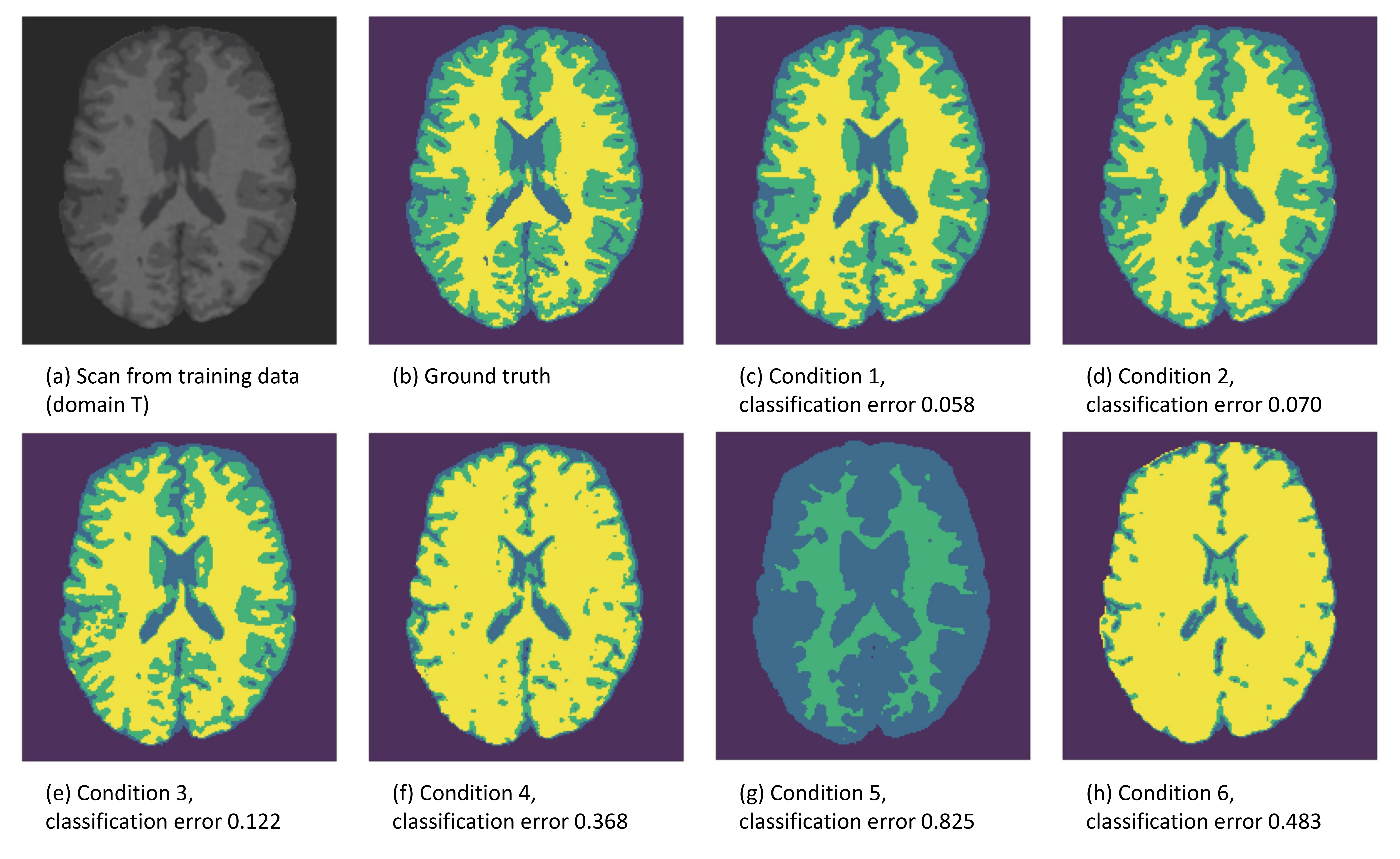}
 	\caption{Images based on predicted tissue classes for all six conditions, where the algorithm was built on training data (patches from 15 scans) and applied to unseen data (1 scan). The classification errors are denoted below the images.}
  	\label{fig:pred_tissue_class}
\end{figure}

\begin{figure}
  \centering	
  \includegraphics[width=0.8\textwidth]{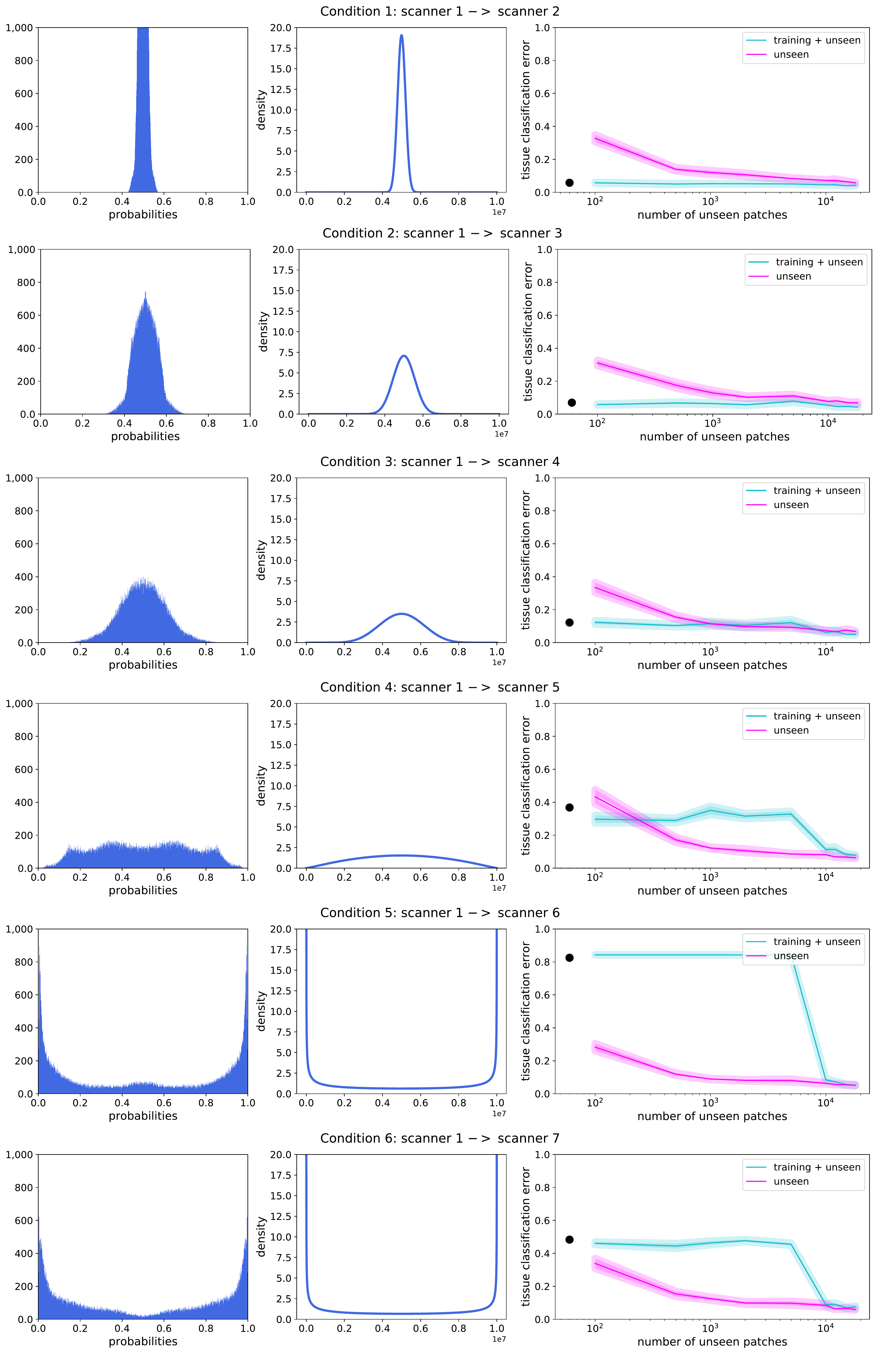}
  \caption{Examples of probability histograms (left column) and corresponding density functions (middle column) are shown for all six conditions, based on the domain classifier. The average (solid line) tissue classification error, along with the standard error of the mean (line thickness) is shown for the training + unseen classifier and the unseen classifier (right column). The tissue classification error is plotted against the number of unseen building patches per scan. The black dots represent the tissue classification error of the rebuilt images as shown in Figure \ref{fig:pred_tissue_class}.}
  \label{fig:tissue_performance}
\end{figure}

Figure \ref{fig:pred_tissue_class} illustrates the effect of data set similarity on the performance of a tissue classifier (built on training data) when applied to a different data set (unseen data), ranging from subtle to severe differences in acquisition parameters between data sets. Results showed that as the difference between the data sets increased, the tissue classification performance decreased dramatically (e.g. conditions 4-6). This is also illustrated in Figure \ref{fig:tissue_performance}, in which the black dots show the tissue classification performance as presented in Figure \ref{fig:pred_tissue_class}. Figure \ref{fig:tissue_performance} further illustrates that as the data similarity grew, the informativeness of the training data set increased.

\begin{table}
\centering
\caption{Tissue classification errors for the six conditions: average with the standard error of the mean between brackets. Errors are given for both the training + unseen classifier and the unseen classifier, for 100, 1,000 and 18,000 unseen building patches per scan.}
\label{tab:tissue_error}
\begin{tabular}{l c c c c c c}
		& \multicolumn{2}{c}{100 unseen patches} & \multicolumn{2}{c}{1,000 unseen patches} & \multicolumn{2}{c}{18,000 unseen patches}  \\
 \Xhline{2\arrayrulewidth}
 & training + unseen & unseen & training + unseen & unseen & training + unseen & unseen \\
 \Xhline{2\arrayrulewidth}
 condition 1 & 0.058 (0.003) & 0.328 (0.021) & 0.052 (0.002) & 0.121 (0.012) & 0.043 (0.003) & 0.057 (0.003) \\ 
 condition 2 & 0.058 (0.003) & 0.311 (0.016) & 0.064 (0.005) & 0.129 (0.016) & 0.042 (0.002) & 0.069 (0.006) \\
 condition 3 & 0.123 (0.011) & 0.335 (0.030) & 0.115 (0.015) & 0.115 (0.005) & 0.005 (0.006) & 0.068 (0.003) \\
 condition 4 & 0.297 (0.024) & 0.433 (0.044) & 0.351 (0.026) & 0.122 (0.006) & 0.078 (0.004) & 0.064 (0.003) \\
 condition 5 & 0.843 (0.001) & 0.283 (0.022) & 0.842 (0.001) & 0.089 (0.004) & 0.051 (0.002) & 0.052 (0.004) \\
 condition 6 & 0.461 (0.008) & 0.339 (0.030) & 0.464 (0.012) & 0.125 (0.009) & 0.075 (0.007) & 0.059 (0.004) \\
 \Xhline{2\arrayrulewidth}
 \end{tabular}
\end{table}

In Figure \ref{fig:tissue_performance} (right column) the tissue classification error is shown for both the training + unseen classifier and the unseen classifier. Recall that the training + unseen classifier was built using both training data and unseen data, while the unseen classifier was built using only unseen data. Both classifiers were tested on unseen data. The tissue classification error is shown as a function of the number of unseen patches per scan for building the model. In Table \ref{tab:tissue_error}, the tissue classification error can be found for 100, 1,000 and 18,000 unseen building patches per scan.

Conditions 1-3 showed a similar tissue classification performance pattern. As the number of unseen patches for building the model increased, the unseen classifier's performance shifted towards the performance of the training + unseen classifier. In condition 3, this shift happened earlier than in conditions 1 and 2. Overall, it was more beneficial to build a classifier on both training and unseen data rather than on merely unseen data, indicating that the training data was informative of the unseen data.

The most interesting finding is seen in condition 4, where we observe a turning point. In this condition, the training + unseen classifier now performed worse than the unseen classifier, indicating that the training data worsened the tissue classification performance. In conditions 5 and 6, the training + unseen classifier also performed worse than the unseen classifier. In such situations, where the data sets were very different, a better classification performance was achieved when only unseen data was used to build the model.

Whether training data is informative of unseen data, can also be seen from domain classification (where the classification only requires domain labels). Recall that the domain classifier was built to distinguish between domain T (training data) and domain U (unseen data). Figure \ref{fig:tissue_performance} (left column) shows the domain classification probabilities, which spread out more as the difference between the training and unseen data increased. Thus, the less similar the domains, the better the domain classifier was able to distinguish between domains. In conditions 1-3, the probabilities were focused around 0.5, indicating that the domain classifier could not distinguish well between the training data and unseen data. This reflects the tissue classification results, where the training data was informative of the unseen data. In condition 4, the probabilities spread out more, where there was no clear focus around 0.5 anymore. The domains started to differ too much, corresponding to the turning point that we observed for the tissue classification. In conditions 5 and 6, the domain classification probabilities were focused around 0 and 1. In these conditions it was easy for the domain classifier to distinguish between domain T and domain U, showing that the training data was not informative of the unseen data.

\subsection{Measuring Data Set Similarity}

\begin{figure}
    \centering
    \includegraphics[width=0.8\textwidth]{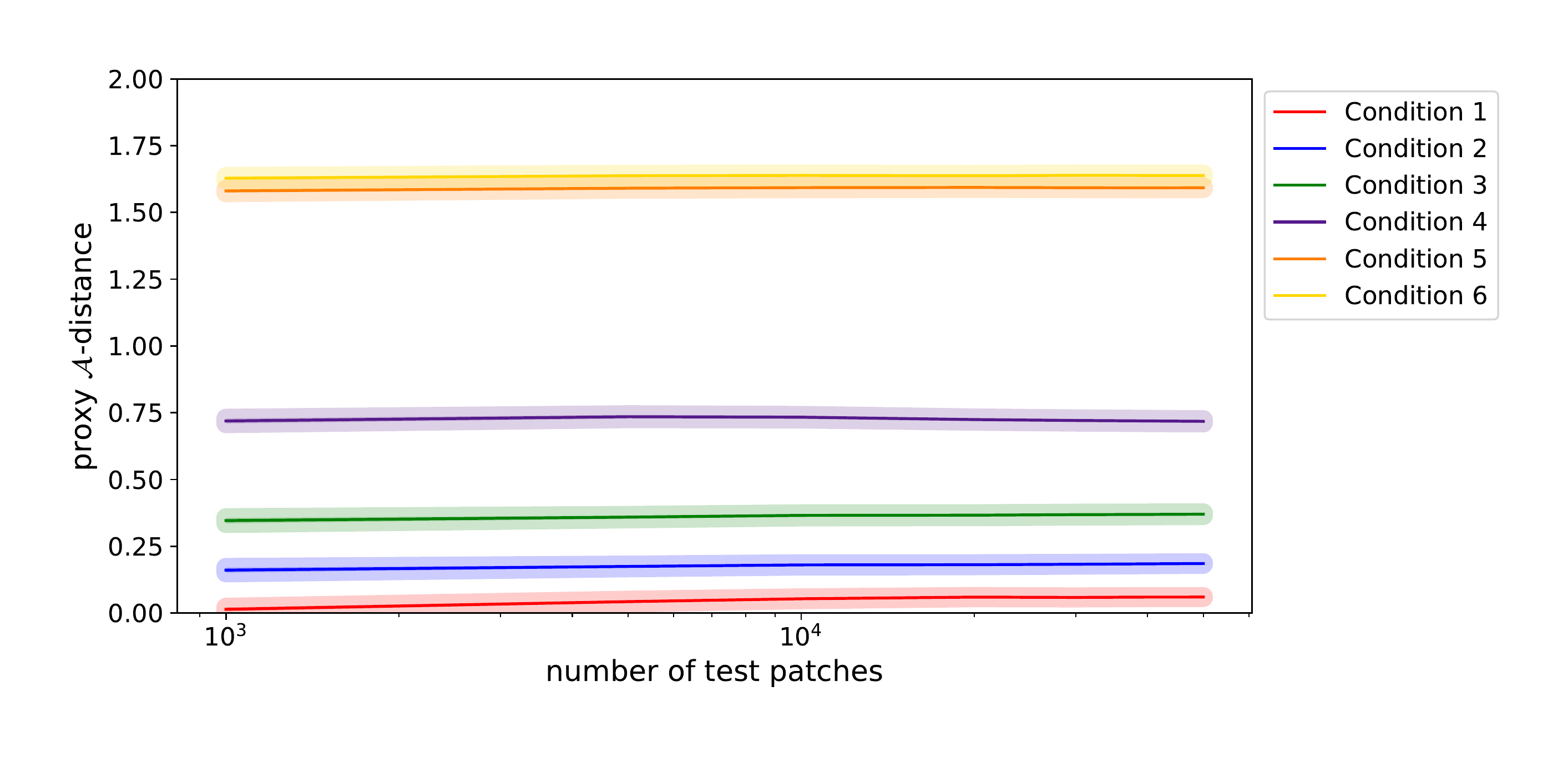}
    \caption{Average proxy ${\cal A}$-distance (solid line) with the standard error of the mean (line thickness) for all six conditions. The proxy ${\cal A}$-distance is plotted against the total number of test patches. Condition 1 provided the lowest proxy ${\cal A}$-distance (largest \textit{similarity} between data sets). Conditions 5 and 6 provided the highest proxy ${\cal A}$-distance (largest \textit{dissimilarity} between data sets).}
    \label{fig:proxyAdistance}
\end{figure}

In the previous section, results showed that as data sets differed more based on domain classification, the training data was less informative for the unseen data (for tissue classification). In this section we present the results of the proxy ${\cal A}$-distance, a measure for data set similarity (i.e. a measure for the left and middle column of Figure \ref{fig:tissue_performance}). 

Figure \ref{fig:proxyAdistance} illustrates that stable predictions for the proxy ${\cal A}$-distance were observed, independent of the number of test patches. The distance between data sets was also represented well, despite it being a simple measure. The high proxy ${\cal A}$-distance for conditions 6 and 7 indicated that the training and unseen data sets were dissimilar, illustrating the large difference in acquisition parameters. On the other hand, the low proxy ${\cal A}$-distance for condition 1 indicated that the training and unseen data sets were very similar, reflecting the small difference in acquisition parameters. As the difference between the data sets became smaller, the proxy ${\cal A}$-distance decreased. The measure was also able to distinguish between subtle differences in conditions. For example, there was a clear difference in proxy ${\cal A}$-distance between conditions 1 and 2. 

\subsection{Data Representativeness Criterion}

\begin{figure}
    \centering
    \includegraphics[width=0.9\textwidth]{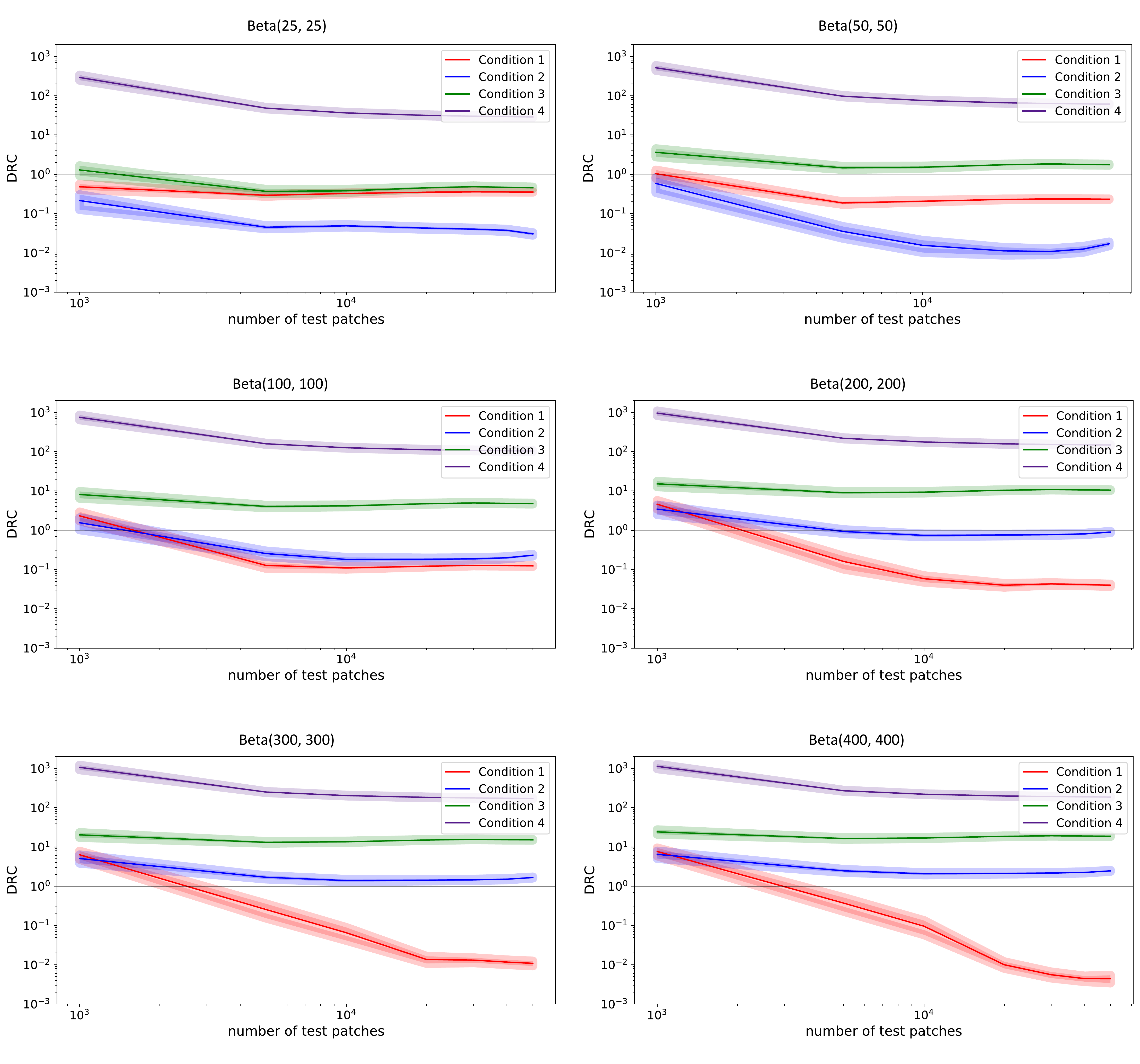}
    \caption{Average DRC (solid line) with the standard error of the mean (line thickness) for conditions 1 to 4, with varying beta shape parameters for benchmark prior 1 as denoted above the plots. Benchmark prior 2 is a Beta(1, 1) distribution. The DRC is plotted against the total number of test patches.}
    \label{fig:DRC}
\end{figure}

In this section we present the results of the DRC. Similar to the proxy ${\cal A}$-distance, the DRC quantifies data set similarity. However, whereas the proxy ${\cal A}$-distance measures the distance between the training and unseen data, it is hard to determine at which point the training data ceases to be representative of the unseen data, which in turn results in a decrease in tissue classification performance. With the DRC, a threshold could be set that determines whether the training data is sufficiently representative of the unseen data.

Figure \ref{fig:DRC} illustrates, for conditions 1-4, how the DRC behaves with different benchmark priors. For conditions 5 and 6, the training data and unseen data were so far apart that the resulting density functions as shown in Figure \ref{fig:tissue_performance} (middle column) were improper. Because of these improper density functions, the DRC could not be acquired.

Figure \ref{fig:DRC} shows that the DRC stabilized with a sufficient amount of patches. The following observations are based on the stabilized DRCs. In all six situations (i.e. different benchmark prior 1), the DRC for condition 1 was always smaller than 1. Similarly, for condition 4 the DRC was always larger than 1. The choice of benchmark prior mostly influenced conditions 2 and 3, where the DRC was either above or below the threshold value of 1 depending on the choice of benchmark prior 1.

Thus, the choice of benchmark prior 1 determines where the DRC of the conditions are with respect to the threshold (i.e. smaller, larger or around 1). By determining the point at which underperformance becomes acceptable, one can determine the strictness of benchmark prior 1. For example, if we relate the DRC to the turning point in tissue classification observed in condition 4 in Figure \ref{fig:tissue_performance}, one could argue to choose a Beta(25, 25) distribution for benchmark prior 1. For the conditions proceeding the observed turning point (i.e. conditions 1-3), where the training data was informative of the unseen data, the DRC was smaller than 1. For condition 4, where the training data was no longer informative of the unseen data, the DRC was larger than 1. 

On the other hand, if the goal is to have a minimum decrease in performance one should choose a very strict benchmark prior. If one considers only conditions 1 and 2 from Figure \ref{fig:pred_tissue_class} to be acceptable, a Beta(100, 100) or Beta(200, 200) distribution would be suitable for benchmark prior 1. If one is a bit more lenient and considers condition 3 to be acceptable as well (similar to relating the DRC to the turning point), a Beta(25, 25) distribution could be chosen. The choice of benchmark prior 1 can be adapted, depending on what one considers an acceptable underperformance. 

\section{Discussion}\label{discussion}
The data representativeness criterion (DRC) determines how representative a training data set is of a new unseen data set. For brain tissue segmentation in MRI data, we showed that the representativeness of the training data as measured by both the proxy ${\cal A}$-distance and the DRC relates to the performance of the supervised tissue classification. Based on the data set similarity, the DRC is able to determine when the performance of the supervised classifier decreases. For a DRC smaller than 1, the training data set can be considered representative of the unseen data set. For a DRC larger than 1, the training data set is \textbf{not} representative of the unseen data. The supervised classification that is based on the training data will therefore under-perform and additional action has to be taken to improve classification performance. Solutions include adding more labeled unseen patches (as shown in proof of principle) or applying representation learning \cite{kouw2019}. If the DRC is around 1, then it is unclear how the algorithm will perform and we recommend proceeding with caution. The strictness of the DRC can be set, depending on the application, using the benchmark prior that determines at which point the underperformance becomes unacceptable. 

As mentioned above, the DRC is based on the similarity between the training data set and the unseen data set. Figure \ref{fig:tissue_performance} shows that as the dissimilarity between the unseen data set and training data set increases, the added value of the training data set decreases. We observed a turning point (Figure \ref{fig:tissue_performance}, condition 4) where the training data is so dissimilar from the unseen data that it is more beneficial to label a small number of patches from the unseen data set to train the supervised classifier, then to add the much larger training data set. This effect increases when data set dissimilarity increases, as shown in Figure \ref{fig:tissue_performance} (conditions 5 and 6). 

Although data set dissimilarity is obvious from a computer vision perspective in Figure \ref{fig:tissue_performance}, it is less obvious from a human vision perspective, when observing the MRI data. Figure \ref{fig:segmentations} shows examples of the simulated MRI scans with different acquisition parameters, ordered based on subtle to severe differences from a computer vision perspective. Conditions 4-6 represent the differences between scans from scanner 1 and scanner 5-7 respectively. Although humans could observe differences between the scans, there is no way of predicting on which scans a supervised classifier would fail, by inspecting these MRI scans. All scans show contrast between white matter, gray matter and cerebrospinal fluid from a human vision perspective. However, Figure \ref{fig:pred_tissue_class} shows that tissue classification could totally fail for conditions 5 and 6, when trained on data from scanner 1. 

One could argue that their dissimilarity could be assessed on the basis of the scanners' acquisition parameters. However, there is no known mapping from specific acquisition parameters to tissue segmentation performance. Furthermore, acquisition parameters are not always known for training data. In this paper we show that determining data set similarity from a computer vision perspective, using the proxy-${\cal A}$ distance and the DRC, has the potential to function as a predictor for supervised classification performance. Other possible applications include determining how representative the training data in machine learning competitions (challenges) is of the test data used to create the leaderboard.

We showed a proof of principle of how the proxy-${\cal A}$ distance and the DRC behave for tissue classification on MRI data. However, there are some limitations that should be taken into account. The DRC could not be computed for all conditions (i.e. scanner comparisons), which restricts the window within which the DRC can be used. It is not possible to obtain the DRC for conditions in which training and unseen data are completely separable (fully dissimilar), such as conditions 5 and 6, as this leads to improper distributions \cite{bousquet2008diagnostics}. The proxy-${\cal A}$ distance, on the other hand, could be determined for all conditions. Condition 5 and 6 show a similar proxy-${\cal A}$ distance, approaching a proxy-${\cal A}$ distance of 2, indicating that the data sets are further apart. However, for conditions 1 to 4 it is unclear when the distance between the data sets is large enough to potentially cause under performance of a supervised classifier. The main added value of the DRC lies in these more subtle cases.

Furthermore, we employed a linear classifier as domain classifier, while there are data sets that are only non-linearly separable. In those cases, a DRC based on a linear classifier could say that data sets are more similar than they actually are. A DRC based on a non-linear classifier, on the other hand, would detect that the data sets are dissimilar. The problem with using non-linear classifiers is that overfitting becomes a much bigger problem. An overfitted non-linear classifier is not reliable either.

This study shows, by means of a proof of principle using MRI data, that the DRC can be used to predict whether a classifier will underperform when applied to a new unseen data set. The DRC can not only be used for the application presented here, but for all applications where one needs to know whether an algorithm built on a training data set will perform sufficiently when applied to a new unseen data set. We argue that the proxy-${\cal A}$ distance is useful in obtaining a general indication as to how similar two data sets are. However, in case the proxy-${\cal A}$ distance is low, one still does not know if a training data set is indeed representative of an unseen data set. Thus, to predict generalization, the DRC should be used.

\section{Conclusion}\label{conclusion}
In this paper we introduced the data representativeness criterion (DRC), to determine whether a training data set is representative of a new unseen data set. For brain tissue segmentation in MRI data, we showed that the representativeness of the training data as measured by both the proxy ${\cal A}$-distance and the DRC relates to the performance of the supervised tissue classification. Based on the data set similarity, the DRC is able to determine when the performance of the supervised classifier decreases. The strictness of the DRC can be set, depending on the application, using the benchmark prior that determines at which point the underperformance becomes unacceptable. The DRC has the potential to be used to predict when additional actions are required, such as adding more labelled data, data augmentation, or representation learning, to improve supervised classification performance on new unseen data sets.  

\section*{Acknowledgements}
RvdS and DV were supported by a grant from the Netherlands organization for scientific research: NWO-VIDI-452-14-006. ES, AM and WK were supported by the Netherlands eScience Center, Research and Development budget.

\bibliographystyle{unsrt}  
\bibliography{references}  






\end{document}